\documentclass[sn-mathphys-num]{sn-jnl}

\usepackage[utf8]{inputenc}
\usepackage[english]{babel}
\usepackage{latexsym}

\usepackage{mathtools}

\usepackage{amsmath}

\usepackage{mathptmx}
\usepackage{enumerate}
\usepackage[labelfont=bf]{caption}[2004/07/16]
\usepackage{color}
\usepackage{verbatim}

\usepackage{mathpartir}
\usepackage{bussproofs}

\usepackage{microtype}
\usepackage[ruled]{algorithm}

\usepackage{graphicx}%
\usepackage{multirow}%
\usepackage{amsmath,amssymb,amsfonts}%
\usepackage{amsthm}%
\usepackage{mathrsfs}%
\usepackage[title]{appendix}%
\usepackage{xcolor}%
\usepackage{textcomp}%
\usepackage{manyfoot}%
\usepackage{booktabs}%
\usepackage{algorithmicx}%
\usepackage{algpseudocode}
\usepackage{listings}%
\usepackage{balance} 
\usepackage{etoolbox}
\usepackage{xurl}
\usepackage{wrapfig}
\usepackage{latexsym,verbatim}
\usepackage{xspace}
\usepackage{xcolor}
\usepackage{enumerate}
\usepackage{array,etoolbox}
\preto\tabular{\setcounter{magicrownumbers}{0}}
\newcounter{magicrownumbers}

\usepackage{makecell}

\NewDocumentCommand{\IfNoValueOrBlank}{ m m }
  { \ifboolexpr {
      test {\IfNoValueTF{#1}} or
      test {\ifblank{#1}} 
    }
    {#2}
    {#1}
  }

\NewDocumentCommand{\MinMax}{ m m }
  { 
    \def\mylower{\IfNoValueOrBlank{#1}{1}}
    \def\myupper{\IfNoValueOrBlank{#2}{n}} 
  }

\NewDocumentCommand{\Ochain}
  { s 
    > { \SplitArgument {1} {:} } O{1:n}
    > { \SplitArgument {1} {_} } m 
  }
  {
    \IfBooleanTF{#1}
    { \myOchain*#2#3 }
    { \myOchain#2#3 }
  }

\NewDocumentCommand{\myOchain}{ s m m m m }
  { \MinMax{#2}{#3}
    \ifboolexpr{
       test {\IfBooleanTF{#1}} and 
       not test {\IfNoValueTF{#5}} }
        { \bigotimes_{#5=\mylower}^{\myupper}#4_{#5} }
        { \bigotimes_{\mylower}^{\myupper}#4\IfNoValueF{#5}{_{#5}} } 
  }

\NewDocumentCommand{\ochain}
  { > { \SplitArgument {1} {:} } O{1:n} m }
  { \myochain#1{#2} }

\NewDocumentCommand{\myochain}{ m m m }
  { 
    \MinMax{#1}{#2} 
    #3_{\mylower}\otimes\dots\otimes #3_{\myupper}
  }

\newcommand{\nmdash}{\Rightarrow}

\makeatother

\algloopdefx{Switch}[1]{\textbf{switch} (#1)}%

\algblockdefx{Case}{EndCase}[1]{\textbf{case} #1\textbf{:}}{Break}

\algcblockdefx{Case}{Case}{EndCase}[1]{\textbf{case} #1\textbf{:}}{\hspace{-0.5cm}\textbf{end switch}}

\algcblockdefx{Case}{Otherwise}{EndOther}{\textbf{otherwise:}}{\hspace{-0.5cm}\textbf{end switch}}

\theoremstyle{thmstyleone}%
\theoremstyle{thmstyletwo}%
\newtheorem{example}{Example}%

\theoremstyle{thmstylethree}%
\newtheorem{definition}{Definition}%

\newtheorem{intuition}{Intuition}

\algloopdefx{Switch}[1]{\textbf{switch} (#1)}%

\newcommand{\PR}{\mathbf{privacy}}
\newcommand{\ND}{\mathbf{non\_discrimination}}
\newcommand{\DG}{\mathbf{dignity}}
\newcommand{\SC}{\mathbf{social\_assistance}}
\newcommand{\HS}{\mathbf{merit}}

\newcommand{\To}{\Rightarrow}

\DeclareMathAlphabet{\mathsf}{T1}{phv}{m}{n}
\DeclareFontFamily{T1}{phv}{\hyphenchar\font=`\-}
\DeclareFontShape{T1}{phv}{m}{n}{<-> s * [.9] phvr8t}{}

\newcommand{\VAL}{\ensuremath{\mathrm{Rights}}\xspace}

\newcommand{\cval}{\ensuremath{\mathbf{right}}\xspace}
\newcommand{\val}[1]{\mathbf{right}^{\mathbf{#1}}}

\newcommand{\cVal}{\ensuremath{\mathbf{R}}\xspace}
\newcommand{\Val}[1]{\mathbf{R}^{\mathbf{#1}}}

\newcommand{\Obl}{\ensuremath{\mathsf{O}}\xspace}

\newcommand{\PROM}{\mathsf{Promotes}}
\newcommand{\DEM}{\mathsf{Demotes}}

\newcommand{\set}[2][\relax]{\ensuremath{#1\{#2#1\}}}
\newcommand{\Set}[2][\relax]{%
  \ensuremath{
    \ifx#1\left
      #1\{#2\right\}
    \else\ifx#1\right
      \left\{#2#1\}
      \else #1\{#2#1\}\fi\fi}%
}
\newcommand{\collide}[3]{\mathsf{Collide}(#1,#2,#3)}
\newcommand{\choice}[2]{\mathsf{Choice}(#1,#2)}

\newcommand{\OBL}{\Obl}

\def\dRule#1:#2=>#3#4{#1\colon #2\To_{#3}#4}

\begin{document}

\title[Foundations for Risk Assessment of AI]{Foundations for Risk Assessment of AI in Protecting Fundamental Rights}


\author[1]{\fnm{Beatrice} \sur{Ferrigno}}\email{beatrice.ferrigno2@unibo.it}

\author*[1]{\fnm{Antonino} \sur{Rotolo}}\email{antonino.rotolo@unibo.it}

\author[1]{\fnm{Jose Miguel Angel} \sur{Garcia Godinez}}\email{miguel.garcia2@unibo.it}

\author[2]{\fnm{Claudio} \sur{Novelli}}\email{claudio.novelli@yale.edu}

\author[1,3]{\fnm{Giovanni} \sur{Sartor}}\email{giovanni.sartor@unibo.it}

\affil[1]{\orgdiv{Department of Legal Studies and Alma AI}, \orgname{University of Bologna}, \orgaddress{\city{Bologna},  \country{Italy}}}

\affil[2]{\orgdiv{Digital Ethics Center}, \orgname{Yale University}, \orgaddress{\city{New Haven},  \state{CT}, \country{USA}}}

\affil[3]{\orgdiv{Department of Law}, \orgname{EUI}, \orgaddress{\city{Florence},  \country{Italy}}}


\abstract{This chapter introduces a conceptual framework for qualitative risk assessment of AI, particularly in the context of the EU AI Act. The framework addresses the complexities of legal compliance and fundamental rights protection by integrating definitional balancing and defeasible reasoning. Definitional balancing employs proportionality analysis to resolve conflicts between competing rights, while defeasible reasoning accommodates the dynamic nature of legal decision-making. Our approach stresses the need for an analysis of AI deployment scenarios and for identifying  potential legal violations and multi-layered impacts on fundamental rights. On the basis of this analysis, we provide philosophical foundations for a logical account of AI risk analysis. In particular, we consider the basic  building blocks for conceptually grasping the interaction between AI  deployment scenarios and fundamental rights, incorporating in defeasible reasoning definitional balancing and arguments about the contextual promotion or demotion of rights. This layered approach allows for more operative models of assessment of both high-risk AI systems and General Purpose AI (GPAI) systems, emphasizing the broader applicability of the latter. Future work aims to develop a formal model and effective algorithms to enhance AI risk assessment, bridging theoretical insights with practical applications to support responsible AI governance.}




\maketitle

\section{Introduction}
\label{sec:introduction}

Legal regulations and governance of AI have become essential topics in literature and research, focusing on aligning AI with human rights, democracy, and the rule of law \cite{Burgess2024}. A variety of strategies for AI governance are being proposed and already adopted, incorporating both international frameworks and a mix of soft and hard law approaches, as well as alternative compliance mechanisms. These strategies include human rights impact assessments, certification and labeling, auditing, regulatory sandboxes for safe AI testing, and automated systems for continuous compliance and risk mitigation. In this context, regulators, AI experts, and representatives from various sectors strive to balance legal frameworks with the fast-paced evolution of AI technology.

Starting from 2023, the global focus on AI policy and regulation intensified, particularly with significant developments in the EU, Africa, and China. After initial moves towards a light but clear regulatory framework for AI, the US now seems to be taking a different policy route; however, the scientific debate in the US remains very lively. In this landscape, the EU has notably introduced the first comprehensive AI Act, emphasizing principles of risk mitigation and transparency.

Existing regulatory approaches to AI can be categorized into three main frameworks: (1) regulation through liability rules, and sanctioning harmful behaviour; while this approach is in principle indisputable, it can be problematic in practice due to the complexity and pervasiveness of AI. Maybe using AI itself is the only effective way to monitor and detect violations (cf. \cite{GeorgeWalsh2022,BalkeCostaPereiraDignumEtAl2013}); (2) legal compliance by design emphasizes the incorporation of legal requirements directly into AI systems, ensuring that they operate based on embedded legal knowledge \cite{Cavoukian2013,Sartor2011,BalkeCostaPereiraDignumEtAl2013}; and (3) compliance as risk mitigation, as outlined in the EU AI Act (AIA), emphasizes adherence to risk management practices. This model suggests that sanctions should be a last resort while proposing alternative mechanisms to prevent risks by inducing compliance with preventive practices. \cite{NovelliEtAl2023,NovelliEtAl2024}. The effectiveness and suitability of these regulatory models in addressing current and future AI challenges remain uncertain.

There are pros and cons for each of the regulatory approaches mentioned above, including the extent to which they can be fully operationalized and managed by competent authorities. This has led to a mixed approach, where there is an increasing demand for developers and deployers to directly consider fundamental rights and values when designing and applying technological solutions for AI. In the context of rapid AI technological development, the law often falls short of fully translating high-level legal and ethical principles into precise prescriptions and compliance methods, making it difficult to ensure top-down implementation. This responsibility is thus delegated to developers and deployers.

However, directly applying fundamental rights and broadly defined social values, such as security, democracy, or the rule of law, poses a formidable challenge. Not only individual citizens but also companies (especially smaller ones) may be unprepared to tackle this. Fundamental rights and values are often undetermined and open to different interpretations and applications. Additionally, tensions frequently arise between different rights or between rights, social values, and economic or political imperatives.

Consequently, developers and deployers need to engage in comprehensive risk and benefit analyses — covering not only health, safety, and environmental sustainability but also fundamental rights and values. They must then assess, according to principles of proportionality, the absolute and relative merit of the technological solutions they propose.

The EU AI Act (AIA) is a very disputed but pivotal step towards a comprehensive legal framework for AI. AIA is a very ambitious legislative project, which revolves around some basic principles summarised below.\\

\begin{intuition}[Basic principles of AIA]
The AIA is grounded on the following basic principles:
\begin{itemize}
 \item Gaps in AI liability have a negative economic impact;
 \item Legal compliance and liability rules can reduce harm related to AI;
 \item Compliance promotes social acceptance of AI;
 \item Provisions must apply to AI providers and to AI deployers;
 \item AI must respect  fundamental rights and the rule of law;
 \item AI systems should be assessed beforehand to protect rights but also to prevent as much as possible litigation in product liability.
\end{itemize}
\end{intuition}

As is well-known, AIA regulates different types of AI systems (AI), covering both general-purpose (GPAIs) and AI specific applications. The focus of the AIA is on high-risk systems \cite{schuett_2023}, which covers domains such as biometric systems, critical infrastructure, education, administration of justice, and democratic processes.  Such systems may pose serious risks but may also provide 
advantages for individuals and the public interest. Thus they must be designed and deployed in such a way that risks are reduced and advantages are preserved as much as possible\cite{NovelliEtAl2023,NovelliEtAl2024}. We should also notice that the AIA acknowledges that certain uses of GPAI systems may count as high-risk and that  GPAI may be integrated into high-risk systems.

The text of AIA leaves open several practical issues and problems to be settled in the implementing acts as well as some broader theoretical questions. In particular, we still need to provide foundations for the concept of legal risk. 

Is risk-based assessment of AI a philosophical challenge in legal reasoning? How can we develop qualitative reasoning models for AI risk assessment? A first preliminary proposal has been advanced by \cite{NovelliGovernatoriRotolo2023}, which made use of the legal-risk framework developed in \cite{NovelliEtAl2023,NovelliEtAl2024}. Complementing the principles behind the AIA, this chapter aims at offering broad conceptual foundations for a high-level assessment method of the impact of AI towards fundamental rights. We will explain how the proposed method can be developed in the contxt of legal reasoning and thus employed for operational compliance tools supporting both the authorities and AI developers and deployers. The chapter elaborates \cite{NovelliGovernatoriRotolo2023,deon2025}'s intuition, aiming at offering a general qualitative reasoning methodology for AI risk.

The layout of the chapter is as follows. Section \ref{sec:fundamental_ rights}
 discusses the concept of legal risk in AI, focusing on unintended adverse consequences. It underscores the multidimensionality of legal risks, highlighting the significant impact on fundamental rights like privacy and non-discrimination. It delineates between material and immaterial harms, and stresses the subjective nature of risk perception among stakeholders. The section frames these challenges within the EU AI Act's regulatory objectives, pointing out potential systemic risks from GPAIs, thus necessitating a proactive regulatory framework. Section \ref{sec:building}
serves as the foundation for understanding AI legal risks. It details the components of normative risk, differentiating between potential violations of legal prescriptions and infringements on fundamental rights and values. To aid risk analysis, a "what-if" approach is proposed, advocating for the examination of layered deployment scenarios. The section also outlines broader risk implications, emphasizing how systemic-risk GPAIs have a wider range of potential impact scenarios and suggesting a necessity for comprehensive evaluation methods. In Section
\ref{sec:balancing_defeasible}
 the integration of definitional balancing and defeasible reasoning is explored as a method for addressing AI legal risks. Definitional balancing is identified as a strategy to resolve rights conflicts by evaluating competing interests, whereas defeasible reasoning provides adaptability in legal decision-making. This section posits that these two building blocks together form a robust framework for resolving conflicts in the AI context, aligning legal compliance with evolving social values and interpretations. Section \ref{sec:conceptual}
outlines a structured methodology for assessing AI deployment scenarios, focusing on the impact on legal rights. It proposes a layered approach progressing from general to specific scenarios. The analysis identifies obligations and rights implicated by each scenario, assesses the impact of AI actions on these rights, and establishes priorities where conflicts arise. The framework is designed to be comprehensive, accommodating both high-risk AI systems and GPAIs. Section \ref{sec:language}
introduces the conceptual pillars to capture and represent interactions between AI deployment scenarios and fundamental rights. The pillars allow for structured reasoning about rights, integrating considerations on promoting and demoting rights within specific contexts. This analysis aims to provide a clear structure to support the qualitative analysis of the impact of AI systems on legal rights, and mechanisms for resolving conflicting rights and interests in a flexible yet structured manner. On this basis of these mechanisms some heuristics derived from the framework are presented to minimize AI legal risks. We discuss the impact of deployment scenarios on rights and suggest practical steps for optimally configuring AI systems or GPAIs to reduce these risks.

\section{Legal Risk of AI: Preliminaries}\label{sec:fundamental_ rights}

The notion of risk points to the unintended adverse consequences of AI technologies. By ``legal risks'' we mean those risks that are relevant to the law, and that therefore have to be prevented, or at least minimized according to proportionality.  The risks to be taken into account according to the AIA include not only threats to health and safety, but also  negative impacts on fundamental rights,  privacy, non-discrimination, and on fundamental values, such as the democracy and the rule of law.
It was recently emphasized that the notion of legal risk---as related to identifying and assessing potential harms that are relevant to the law---is multi-dimensional \cite{Kusche2024}. This especially holds with
regard to the notion of harm to fundamental rights. In particular, following \cite{Kusche2024} we should observe:

\begin{itemize}
 \item{\bf Immaterial vs. material harms:} While material harms cover physical or financial damage, immaterial harms include infringements on privacy, mental health, social equality, and ethical standards. The EU AI Act highlights the importance of addressing both, particularly as AI systems increasingly affect non-quantifiable aspects of life like dignity and autonomy.
\item{\bf Perspectives on AI risk:}
 Stemming from sociological risk research, which emphasizes the subjective nature of risk perception, AI risks are often viewed differently by various stakeholders. Understanding this multiplicity of perspectives is crucial in constructing a regulatory framework that accommodates diverse concerns and expectations.
\item{\bf Systems theory and risk perception} Drawing on systems-theoretical approaches, one notes that risk perception varies significantly across social actors. The regulation must thus seek a common ground that aligns legal standards with ethical considerations. This involves balancing individual rights with societal needs, a complex task given AI's wide-ranging effects.
\item{\bf Complexities and tensions in regulation: }
The EU AI Act attempts to mitigate risks through a structured regulatory approach but faces challenges inherent in risk assessment and control. The Act's reliance on fundamental rights as both a safeguard and a vulnerable point creates tensions within regulatory practices.
\item{\bf Fundamental rights and political values:} Fundamental rights (e.g., freedom of opinion or privacy) are both legal constructs and political values. Thus, they need to be dynamically interpreted, according to current societal values. This flexibility can be beneficial but also problematic when delineating clear boundaries for what's considered safe or ethical AI.
\item{\bf Trustworthiness and compliance: } The Act promotes the concept of trustworthy AI, suggesting compliance with regulations as a proxy for trust. However, trust cannot be legislated – it requires ongoing proof of alignment with ethical standards and societal expectations. There is a  tension between regulating to foster trust while recognizing that trust is culturally and socially constructed.
\item{\bf Addressing systemic risks: } The onset of GPAI models poses systemic risks that transcend individual sectors or regions. These models, capable of performing across various domains, introduce complexities that necessitate adaptive regulatory frameworks to address potential disruptions within democratic processes, economic stability, and social equity.
\end{itemize}

\section{AI Legal Risk: Building Blocks}\label{sec:building}

Legal risks pertain to the potential impact of AI applications on legal standards and fundamental rights. By exploring these risks, we can better understand the implications of deploying AI systems and develop appropriate safeguards to mitigate adverse effects.

\subsection{Normative Risk: Legal Violations and Fundamental Rights}

At the core of legal risk is the consideration of how AI might infringe upon legal prescriptions and fundamental rights, as outlined in the AI Act.\\

\begin{enumerate}
    \item \textbf{Potential Violation of Legal Prescriptions}: GPAIs may breach existing laws in various deployment scenarios, including sensitive sectors—such as finance or healthcare—must governed by stringent regulations regarding data protection, discrimination, and transparency. The inability of these models to contextualize legal constraints could lead to inadvertent violations.
    
    \item \textbf{Infringement of Fundamental Rights and Values}: AI systems can impact on individual liberties, such as the right to privacy and non-discrimination. When GPAIs are integrated into algorithms that influence critical life decisions—such as employment opportunities or healthcare access—the risk of bias and unfair treatment increases. These systems must be scrutinized to prevent them from exacerbating societal inequalities or infringing on personal rights.\\
\end{enumerate}

The following intuition is thus a basic building block:\\

\begin{intuition}[AI legal risk: two dimensions]\label{intuition:two}
Two interrelated aspects must be considered:
\begin{enumerate}
 \item assessing the potential impact of AIs on the compliance with obligations applying to them;
 \item assessing the potential detriment of fundamental rights determined by the deployment of AIs.
\end{enumerate}
\end{intuition}

\subsection{Normative Risk Analysis: A What-If-Analysis Approach}

To conduct a robust legal risk analysis, it is essential to adopt a what-if approach, examining a range of deployment scenarios across multiple layers.

\begin{itemize}
    \item \textbf{High-level scenarios}: At the top tier, we can outline abstract deployment scenarios where AI systems operate within generalized environments. These generalized contexts provide a foundational understanding but lack specificity regarding application consequences.
    
    \item \textbf{Layered deployment scenarios}: As we move down the hierarchy, these high-level scenarios can be articulated into more detailed applications. For instance, a high-level scenario might involve the use of AI in law enforcement, while more specific scenarios would involve the application of AI for facial recognition, predictive policing, or evidence evaluation. Each specific scenario can then be examined for its potential normative impacts, providing insights into possible infringements of rights and legal violations.
\end{itemize}

This layered analysis allows stakeholders—including regulators, developers, and users—to appreciate the nuanced risks associated with various deployment contexts, facilitating informed decision-making in AI governance.

The following intuition is thus another basic building block:\\

\begin{intuition}[AI legal risk: two dimensions]\label{intuition:what-if}
For any AI system or model $\sigma$, identifying  legal risks, even at the development and AI-provider level, requires a what-if analysis to:
\begin{enumerate}
 \item determine the characterizing deployment scenarios $S_1, \dots  , S_n$ of $\sigma$;
 \item assess the compliance impact of $\sigma$ as outlined in Intuition \ref{intuition:two}.
\end{enumerate}
\end{intuition}

\subsection{Broader Risk Implications: The Superset of Deployment Scenarios}

Building on the earlier points, the relationship between systemic-risk GPAIs and high-risk AI systems illustrates that the broader the perceived risk, the more deployment scenarios must be examined.

\begin{itemize}
    \item \textbf{Superset of scenarios}: The set of high-level deployment scenarios pertaining to systemic-risk GPAIs is indeed a superset of those concerning high-risk AI systems. This means that while not all GPAI applications will inherently infringe on rights or legal norms, the potential for such violations exists across diverse deployment scenarios, particularly in multifaceted applications integrating various AI technologies.
    
    \item \textbf{Families of deployment scenarios}: Each high-risk AI category encompasses a set of deployment scenarios that warrant distinct consideration. For instance, in healthcare, deploying AI for diagnosis or treatment planning accompanies unique normative risks tied to patient rights, accuracy of information, and potential biases in treatment recommendations.
\end{itemize}

Consequently, normative risk evaluation within the scope of the AI Act must be comprehensive, accounting for the multiplicity of deployment scenarios associated with varying levels of risk. 

\begin{figure}[htb]
\begin{center}
    \includegraphics[width=.75\columnwidth]{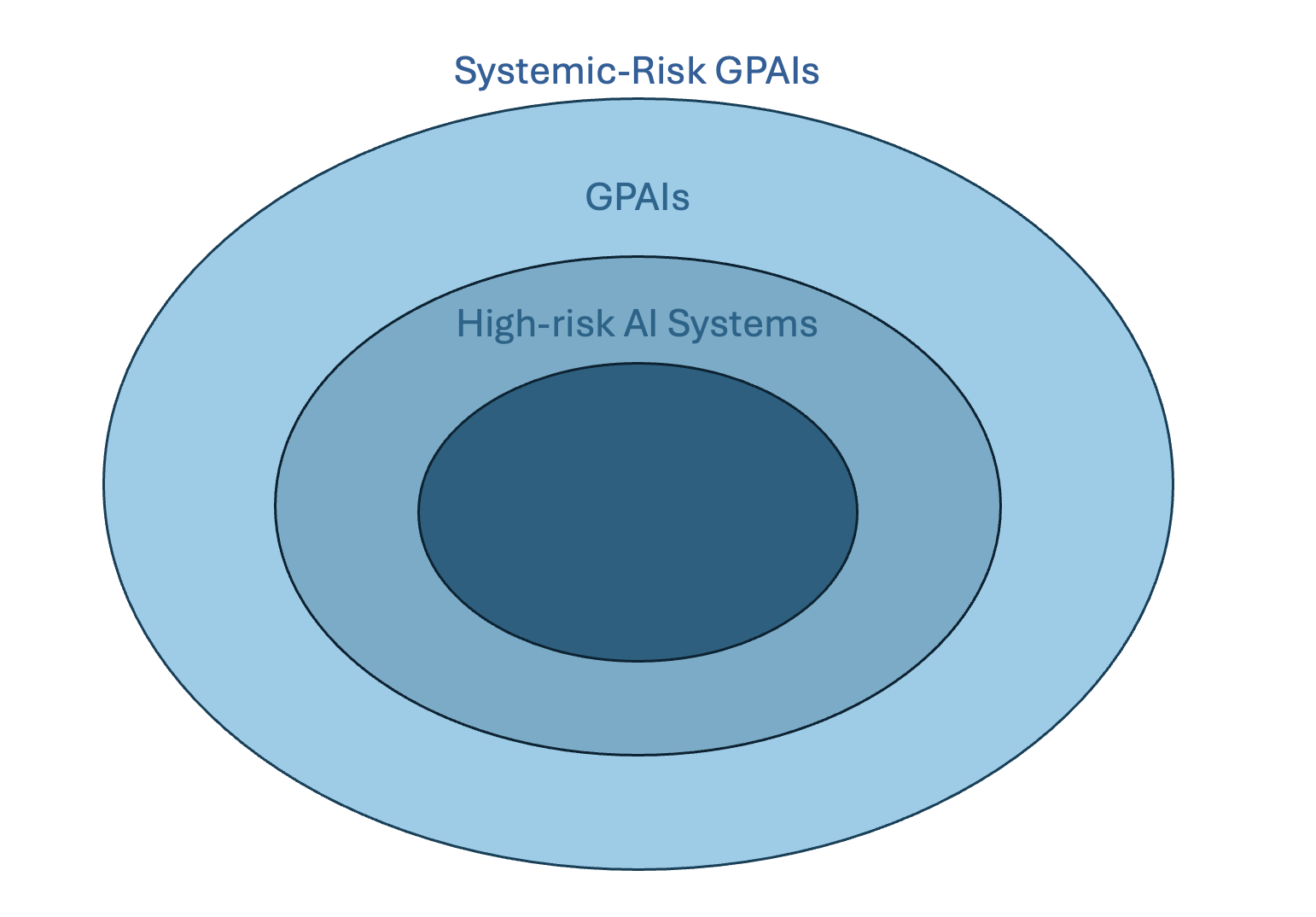}
    \caption{Relation among AI types regarding  characterizing deployment scenarios}
  \label{fig:scheme_deployment}
  \end{center}
\end{figure} 

\subsection{An Example for AI Deployment: The Fundamental Rights Impact Assessment}

The AIA, at Article 26, introduces new transparency obligations emphasizing the protection of individuals' fundamental rights in AI decision-making. Deployers must notify individuals when these decision-making systems are applied, clarify their intended purpose, and specify the types of decisions involved. Article 27 introduces a \emph{Fundamental Rights Impact Assessment} (FRIA) before deploying high-risk AI systems. The FRIA mandates that  deployers of high-risk systems that are bodies governed by public law or  private operators providing public services, as well as all  deployers of certain kinds of systems shall perform an
assessment of the impact on
fundamental rights that the use of
the system may produce. This assessment should encompass factors such as intended purpose, scope, impact on marginalized groups, environmental considerations, compliance with legal and fundamental rights, and democratic implications for public authorities.

Deployers must develop a detailed plan to mitigate negative fundamental rights impacts or inform the AI provider and national authorities promptly. Relevant stakeholders, including equality bodies, consumer protection agencies, and data protection authorities, should be involved, with a six-week window for their input. Public bodies must publish the assessment as part of their EU registration.

The FRIA is framed in the context of art. 26, which states the following obligations for  deployers:
    \begin{itemize}
        \item Take appropriate technical and organizational measures to ensure the use of systems in accordance with accompanying instructions.
        \item Ensure competence, training, and authority for personnel overseeing high-risk AI systems.
        \item Ensure input data relevance and representativeness.
        \item Monitor system operation, inform providers of risks and suspend use if necessary.
        \item Maintain logs for a specified period and inform relevant authorities of incidents.
        \item Inform workers representatives and affected workers before deploying AI systems at the workplace.
        \item Comply with registration obligations for public authorities or EU institutions.
        \item Use provided information to conduct data protection impact assessments where applicable.
        \item Obtain authorization for the use of AI systems for criminal investigation purposes.
        \item Document each use of AI systems for law enforcement and submit annual reports to relevant authorities.
        \item Inform natural persons subject to AI system use and cooperate with national authorities to implement regulations.
    \end{itemize}

The core idea in AIA behind the provision for the FRIA in Art. 27 relies on an ``assessment of the impact on fundamental rights that the use of the system may produce'',  which requires (in addition to specifying the period and frequency in which the AI system is intended to be used, and the subjects involved)
\begin{quote}
\begin{description}
\item[(a)] a description of the deployer's processes in which the high-risk AI system will be
used in line with its intended purpose; [\dots]
\item[(d)] the specific risks of harm likely to impact the categories of persons or groups of
persons identified under point (c), taking into account the information given by
the provider [\dots];
\item[(e)] a description of the implementation of human oversight measures, according to the instructions of use;
\item[(f)] the measures to be taken in case of the materialization of these risks, including their
arrangements for internal governance and complaint mechanisms.
\end{description}
\end{quote}

The crucial provision is par.(d), which conceptualizes the AI risk assessment as focusing on the type of risk related to the potential harm caused by AI systems that may infringe upon legal values or fundamental rights \cite{Kusche2024}.


According to our discussion in Section \ref{sec:building}, the specification of legal risks should be grounded on deployment scenarios $S_1, \dots , S_n$, i.e., on contexts and processes in which the AI system will be
used (art. 27, par. (a)).
Once this is done, one can argue that the potential harm to fundamental rights must be assessed by evaluating the risk that each $S_j$, $1\leq j \leq n$, lead to violating any legal obligation applying to $S_j$. In this sense, the assessment of impact on fundamental rights is only \emph{indirectly} implemented. 

While the interpretation above may cover many aspects of the risk assessment in AI deployment, we argued in \cite{NovelliEtAl2023,NovelliEtAl2024} that a \emph{direct} impact on fundamental rights is possible, provided that an appropriate approach to legal risk is adopted. 

\section{Risk Assessment and the Direct Impact of AI on Fundamental Rights}\label{sec:IPPC}
It was argued in \cite{NovelliEtAl2023,NovelliEtAl2024} that by classifying systems as high-risk according to their intended use, the AI act may predetermine outcomes in the balancing test of rights and interests, without flexibility for risk management adjustments based on changing circumstances.  This can result in an inaccurate evaluation of AI risk, leading to either overly strict or lenient regulations as well as preventive measures and remedies. 
\cite{NovelliEtAl2023,NovelliEtAl2024} have argued that, when evaluating the significant risk and the impact on fundamental rights, we have to shift from a purely scope-oriented categorisation  to an analysis based on risk scenarios involving interactions among multiple risk factors.

To do so, it was proposed to adapt in the context of AIs the risk assessment model arising from the Intergovernmental Panel on Climate Change (IPCC) and related literature. This integrated model enables the estimation of AI risk magnitude by considering the interaction between (a) risk determinants, (b) individual drivers of determinants, and (c) multiple risk types \cite{AI-society}. Risk determinants are the consequences of (a1) hazard, (a2) exposure, (a3) vulnerability, and (a4) response. Hazard refers to potential sources of harm. Exposure refers to what might be affected by the hazard source. Vulnerability refers to attributes or circumstances that make exposed elements susceptible to harm. Response refers to existing measures that counteract or mitigate risk. 

Consider the following example.\\

\begin{example}\label{ex:IPCC}
Consider an AI system for recidivism rate assessment in justice. The following aspcts can be identified:

\begin{description}
\item[{\bf Hazards.}] Causes of adverse effects, e.g.: inner opacity and the poor quality or misuse of the training data. When hazard drivers compound, they can lead to discrimination biases.
\item[Exposure.] Adverse effects, according to substantive legal principles – e.g.,  criminal culpability and equality – and some procedural ones – e.g.,   transparency and the right of/to defense.
\item[\bf Vulnerability.] Attributes that may make individuals or groups susceptible to the adverse effects of AI such as ethnicity, economic conditions, and education.
\item[Response.] Measures that counter the hazards, such as governance or deployment measures.
\item[\bf Extrinsic risks.] External conditions that contribute to the risk such as  compliance risk, liability risk, and economic risk, e.g., lack of effective rules for the allocation of liabilities for adverse effects or loss of opportunities and digital sovereignty.
\end{description}
\end{example}

The above configuration of determinants must be considered to assess the legal risk of an AI system or a GPAI:\\
\begin{itemize}
\item{\bf Step 1: Determining Risk Magnitude:} The overall risk magnitude is determined by two interacting variables: likelihood and severity.
\begin{itemize}
    \item \textbf{Likelihood:} The likelihood of a detrimental event occurring depends on the interplay between hazard drivers (factors increasing risk) and response drivers (factors mitigating risk, such as preventive measures).  A higher prevalence of hazard drivers and/or a lower efficacy of response drivers will increase the likelihood.
    \item \textbf{Severity:} The severity of the detriment resulting from an event is influenced by several factors: the nature and intensity of the hazard sources, the characteristics of the exposed asset (the AI system and its applications), and the vulnerability profiles of the system and its users.  Higher hazard intensity, greater asset sensitivity, and increased vulnerability contribute to greater severity.
\end{itemize}
The interaction between these factors determines the overall risk magnitude.  For instance, a high-likelihood event with low severity may pose a different overall risk than a low-likelihood event with high severity.
\item{\bf Step 2: Suitability Assessment via Balancing Test:}
After quantifying risk magnitude (Step 1), a suitability assessment is crucial. This involves a balancing test to evaluate the acceptability of the identified risk level in relation to the asset exposed (the specific AI application and its context). The balancing test weighs the benefits of the AI application against the potential detriments, considering the magnitude of risk determined in Step 1.  If the benefits significantly outweigh the potential detriments (even considering the risk magnitude), the application may be deemed suitable despite the inherent risk.  Conversely, if the detriments significantly outweigh the benefits, despite mitigation efforts, the application may be deemed unsuitable.  This is a qualitative judgment, requiring careful consideration of the specific context and stakeholders' values.
\end{itemize}



How to reconstruct this balancing process within the what-if analysis mentioned above?


%
\section{Balancing and Defeasible Reasoning for AI Legal Risk}\label{sec:balancing_defeasible}

\subsection{Basic Concepts}

We need to distinguish two kind of balancing: definition and case-by-case balancing. In Definitional balancing the decision-maker besides advancing a solution that balances the conflicts principles in the case at hand, also provides a rule, potentially applicable to future cases stating  conditions under which this solution is appropriate; in case-by-case-balancing the decision-maker only states the solution for the case at hand.  ]
\textit{Definitional balancing} has been used in constitutional law used to resolve conflicts between competing rights or interests by evaluating their relative importance or defining their normative boundaries within specific decision-making contexts. Unlike \textit{ad hoc} (or case-by-case) balancing, definitional balancing enables constitutional courts to establish general principles or precedents, offering guidance to lower courts and legislatures in predicting normative outcomes for future cases.\footnote{See \cite[Chapter~1]{giovanella2017copyright} for a discussion about both definitional and case-by-case balancing.} Since the goal of balancing is to resolve conflicts in a fair, rational, and proportionate way, it aligns with the fundamental principles of constitutionalism and the rule of law, making it a suitable approach to justify decisions affecting key EU values (e.g., security, public health, and democracy).

A fundamental premise of definitional balancing is that rights are not absolute but must be weighed against each other in conflict situations. This is something that requires \textit{proportionality analysis}, which involves a structured evaluation of several factors \cite{Alexy2014-ALECRA-2}. The process begins by assessing whether the goal pursued by limiting a right is \textit{legitimate}, then considers whether the measures adopted to achieve that goal are both \textit{suitable} and \textit{necessary}. Finally, the analysis ensures that the benefits of the limitation outweigh the harm caused to the right, achieving an \textit{equilibrium} between the corresponding competing values. This method thus enables a nuanced and context-sensitive understanding of the scope and limitations of rights within a legal system.

In essence, definitional balancing examines the conceptual and normative interplay between rights and their limitations. For instance, a law restricting free speech might be justified to prevent hate speech, but such justification demands careful delineation of both rights’ boundaries. Balancing, therefore, becomes indispensable for defining how conflicting rights coexist within a legal framework while preserving their normative significance. As \cite{10.1093/icon/mom023} observes, this approach identifies the sources of conflicts and sets the conditions for their resolution, which ultimately amounts to determining the \textit{legal possibilities} of their joint realization. Thus, under this characterization, fundamental rights function as guiding norms that, while subject to exceptions and further revisions, must adapt to the complexities of specific circumstances to maintain their practical relevance.\footnote{Though, discussion remains as to whether rights can be understood not only as principles but rather as \textit{hybrids} of principles and rules (i.e., rules that incorporate the balancing process required to determine the normative content of the principle in the concrete set of circumstances). On this, see \cite[p.~457]{10.1093/icon/mom023}.}
  
\textit{Defeasible reasoning}, on the other hand, is a method designed to handle situations where conclusions can be overturned by adding information (e.g., new evidence) to or revising preference orders (e.g., priority relations) within the original set of premises \cite{pollock1987defeasible} and \cite{morgan2000nature} or arguments \cite{Sartor2018DL}. This approach is characterized by its \textit{adaptability}, capturing the dynamic and nuanced nature of real-life decision-making. Unlike classical logic, which operates rigidly and irrevocably within fixed-point frameworks, defeasible reasoning embraces \textit{non-monotonicity} (allowing conclusions drawn from existing premises to be adjusted or withdrawn in light of new information or shift of preferences) \cite{reiter1988nonmonotonic}. This adaptability makes it particularly well-suited to legal reasoning, where evolving interpretations and circumstances can reshape fundamental normative values.

A defining feature of defeasible reasoning is its recognition of \textit{exceptions}. By introducing a conceptual and logical distinction between broadly applicable rules and specific defeaters, defeasible reasoning accommodates the complexity of situations where general principles must yield to unique conditions \cite{horty2012reasons}.\footnote{For a more elaborated discussion of defeaters in epistemic and practical reasoning, see \cite{bagnoli2018defeaters}.} Within this framework, then, defeasible rules act as normative guidelines, while defeaters establish the boundaries of their applicability. This ensures that decisions remain accurate and context-sensitive.

Furthermore, defeasible reasoning operates as a mechanism to settle conflicts between competing defeasible rules by introducing \textit{defeasible} priority relations \cite{governatori2019revision}. With the organization of these rules into priority structures, defeasible reasoning ensures consistency while retaining the flexibility to adapt to uncertain or evolving circumstances. Together, these elements form a robust system for navigating the complexities of reasoning within practical and epistemic domains.

\subsection{Development}

Balancing competing rights is among the most challenging aspects of legal reasoning, demanding a flexible and nuanced decision-making framework. The integration of definitional balancing and defeasible reasoning provides a comprehensive tool for addressing conflicts involving fundamental rights, particularly in the context of the EU AI Act and legal risk assessment. This synthesis draws upon these theoretical insights (or \textit{building blocks}) to capture the dynamic nature of conflicts of rights while maintaining a structured approach to compliance and adjudication.

This integrated framework recognizes that fundamental rights are not absolute but may be subject to exceptions informed by further considerations. By blending the non-monotonic character of defeasible reasoning with the normative depth of definitional balancing, this framework offers a dynamic and robust mechanism for analyzing and resolving conflicts between fundamental rights in specific AI scenarios. Building on the theoretical convergence of these concepts, this approach takes initial steps toward addressing the multifaceted and evolving challenges of AI governance.

At its core, definitional balancing mirrors the logic of defeasibility. Fundamental rights are treated as defeasible rules, with their limitations functioning as defeaters (i.e., conditions under which a right may be overridden). Proportionality analysis, central to balancing, becomes the mechanism for resolving these conflicts. For example, the right to freedom of speech (a defeasible rule) may be curtailed to prevent incitement to violence (a defeater, justified through proportionality). Similarly, in cases involving privacy and freedom of expression, initial conclusions favoring one right may shift as new facts or priorities emerge. This shared emphasis on non-monotonic reasoning ensures that decisions remain flexible and responsive to changing contexts or circumstances.

This adaptability is crucial for legal compliance and adjudication, where context-sensitive information often determines outcomes. Defeasible logic formalises this flexibility by encoding rules that \textit{activate} or \textit{deactivate} based on the obtaining of certain relevant information. For instance, privacy rights may be temporarily diminished for national security reasons but restored once the threat subsides. This precision is essential for resolving complex rights conflicts where rigid, one-size-fits-all solutions are inadequate.

We argue that this framework is useful in the ethical and legal evaluation of AI technologies. Under the EU AI Act, we need to assess potential impacts on privacy, transparency, non-discrimination, etc. In this context, definitional balancing and defeasible reasoning work in tandem: rights such as privacy and non-discrimination serve as defeasible rules, while public safety or efficiency concerns act as defeaters. For example, deploying facial recognition for law enforcement raises significant privacy concerns. However, under certain conditions (e.g., demonstrable public safety benefits and no less intrusive alternatives), such limitations might be justified through proportionality analysis. Safeguards like bias audits and oversight are essential to meet proportionality standards; without them, privacy and equality concerns prevail.

Defeasible reasoning further enriches this process through its capacity for dynamic conflict resolution. As new evidence emerges (e.g., data revealing systemic bias in an AI system) the justification for its deployment can be defeated, prompting \textit{recalibration} or \textit{discontinuation}. This iterative approach ensures decisions remain grounded in the most up-to-date information.

This framework is based also on solid philosophical principles, which support its practical applications. It aligns with \textit{legal positivism}, recognizing that rights must be interpreted within evolving social and legal contexts rather than taking them as fixed moral truths. \textit{Pragmatism} underscores its focus on common-sense solutions, acknowledging that objectivity is an ideal rather than a strict condition. And \textit{pluralism} regarding theories of justice emphasises the importance of balancing competing claims to prevent any single right from dominating at the expense of others.

We can illustrate this integration with a case study involving biometric surveillance in public spaces. When a state agency proposes deploying AI-based facial recognition to enhance security, rights to privacy and non-discrimination clash with the public interest in safety. Privacy (a default rule) asserts the individual's right to be free from intrusive monitoring. Public safety (another default rule) introduces conditions under which privacy might be limited. However, a certain level of discrimination (e.g., targeting minorities) acts as a further defeater, reinforcing the need for a balanced approach. Proportionality analysis evaluates whether the system is suitable for reducing crime, necessary compared to less intrusive alternatives and whether its benefits outweigh its risks. With appropriate safeguards, the system may pass this test; without them, concerns about privacy and equality dominate.

The integration of definitional balancing and defeasible reasoning, then, offers a robust yet flexible framework for demarcating the normative scope of fundamental rights in specific situations (where both normative and factual information is subject to constant revision). This framework also provides the basis for harmonizing abstract principles with practical adaptability, ensuring that legal and ethical evaluation remains context-sensitive and responsive to societal needs.

\section{Principles of Legal Reasoning for the Impact of AI towards Fundamental Rights}
\label{sec:language}
The conceptual analysis of Section \ref{sec:building}  provides the building blocks for a high-level legal reasoning model for normative risk analysis within the AI Act. In the remainder of this chapter we use logic for an informal outline of a method for modeling normative risk assessment \emph{focused on assessing rights impact}, i.e., the second layer among those discussed in Section  \ref{sec:building}: assessing the potential detriment of fundamental rights determined by the deployment of AIs. 

This second layer, which is the most problematic and for which we argued that a proportionality judgment must be modeled to check, e.g., if the relative negative impact of an AI with respect to legal value $v_1$ is balanced by the promotion of another value $v_2$ \cite{sartor}.

Our approach presents a structured high-level qualitative methodology for analyzing the legal risks associated with deploying AI systems, focusing on both high-risk AI systems and general-purpose AI (GPAI) systems.  The analysis is crucial for ensuring compliance with legal obligations and protecting fundamental rights.  The approach builds upon the core idea that AI risk assessment needs to consider both potential legal violations and the impact on fundamental rights (Intuition \ref{intuition:two}).  Furthermore, a what-if-analysis approach (Intuition \ref{intuition:what-if}) is employed, examining a range of deployment scenarios to anticipate potential problems.

\subsection{The Structure of Fundamental Rights}\label{sec:structure}
We cannot develop any model of AI risk assessment concerning fundamental rights without first analyzing those rights. Many theories have been developed for examining the logical structure of fundamental rights (see an overview of different aspects in \cite{handbook:deontic,handbook:deontic_marek,handbook:deontic_jones,Sartor05}). In general, fundamental rights can be considered as Boolean combinations of basic rights, obligations characterizing their protection \cite{ferrajoli}, and, following Hohfeld's theory \cite{Hohfeld}, powers \cite{Sartor05,GelatiRSG04}.

For simplicity, here we assume that fundamental rights are composed solely of basic rights. Given a predetermined set of basic rights $R = \set{\val{1}, \dots, \val{k}}$, we can construct (in general, but specifically for each deployment scenario) a set of relevant fundamental rights $\set{\Val{1}, \dots, \Val{n}}$.\\

\begin{example}
Consider the right to privacy ($\mathbf{privacy}$) The right to privacy encompasses several components, which might include, e.g., the following components:
\begin{itemize}
   \item Right to data protection ($\mathbf{data\_protection}$): Ensuring personal information is collected and used responsibly;
   \item Right to personal autonomy ($\mathbf{autonomy}$): Allowing individuals to make decisions about their private life without interference;
   \item Right to confidentiality ($\mathbf{confidentiality}$): Protecting personal communications from being accessed or disclosed by others without consent;
    \item Right to dignity ($\mathbf{dignity}$): Ensuring that individuals are treated with respect and that their personal space and intimate relations are protected;
    \item Right to control personal information ($\mathbf{control}$): Allowing individuals to choose what personal data they wish to share and with whom.
\end{itemize}
This means that $\mathbf{privacy}$ corresponds to the following conjunctive combination of basic rights:
\begin{gather*}
\mathbf{privacy} \leftrightarrow (\mathbf{data\_protection} \wedge \mathbf{autonomy} \wedge \mathbf{confidentiality} \wedge \mathbf{dignity} \wedge \mathbf{control}).
\end{gather*}
\end{example}

\subsection{Right Collision}\label{sec:collision}
As we have argued in Sections \ref{sec:IPPC} and \ref{sec:balancing_defeasible} risk assessment requires some form of balancing, because the deployment of AI in any scenario can pose a normative collision among some of the involved fundamental rights (for a general account, \cite{Alexy2014-ALECRA-2,Alexy2009}).

Given the analysis of Section \ref{sec:structure} above, two rights 
$\Val{i}$ and $\Val{j}$ trivially collide if they are logically incompatible, i.e., if $\Val{i} \leftrightarrow \neg \Val{j}$.
%
%
But rights collision goes beyond simple logical incompatibility. Moreover, in addition to standard consistency of rights sentences, we need to express when rights potentially and normatively collide in an AI deployment scenario $X$: in fact, this type of normative collision is contextual \cite{Alexy2014-ALECRA-2,Alexy2009}. If $\bigwedge_{k=0}^{z} f_{k}$ is a logical description of $X$, we can express the collision between $\Val{i}$ and $\Val{j}$ as $\collide{\Val{i}}{\Val{j}}{\bigwedge_{k=0}^{z} f_{k}}$. 

\emph{It is important to observe that the same fundamental rights can be compatible in some AI deployment scenarios and can collide in others}.

\subsection{Rights Promotion and Demotion}\label{sec:promotion}
An AI deployment scenario can promote or demote a certain fundamental right. That a right $\Val{i}$ is promoted by $X$ means that $X$ supports the realization and enhancement of $\Val{i}$; accordingly, if a right $\Val{j}$ is demoted by $X$, this means that $X$ prevents the realization and enhancement of $\Val{j}$. Hence, from any contexts $X, Y$ and fundamental right $\Val{i}$ we write $\PROM(X,\Val{i})$ and $\DEM(Y,\Val{i})$ to denote, respectively, that $X$ promotes or demotes $\Val{i}$.\\

\begin{example}[AI scholarship allocation]
Consider a deployment scenario where the intended purpose of an AI system is supporting an academic institution to allocate scholarships to university students. 
We may consider three deployment scenarios, one where all data protection requirements are ensured and where the student consent is needed to use the AI system, one in which the consent is not adopted but the AI is fully transparent and the criterion for allocating scholarships is only student's low income, and a third one like the second one but where the only criterion is academic performance.

\begin{align*}
S_d = & \{\mathit{student\_consent}, \mathit{min\_datastorage}\}\\
\qquad S_r = & \{\neg \mathit{student\_consent}, \mathit{transparency}, \mathit{student\_income}\}\\
S_e = & \{\neg \mathit{student\_consent}, \mathit{transparency}, \mathit{student\_CV} \}
\end{align*}

The rights involved are:
\[
\{\PR, \ND, \DG, \SC, \HS\}
\]

The different scenarios behave differently regarding which rights are promoted or demoted:
\[
\begin{aligned}[t]
\phi_1 &= \PROM(S_d,\PR)      & \phi_4 &= \PROM(S_r,\SC)\\
\phi_2 &= \PROM(S_d,\ND)   & \phi_5 &= \PROM(S_e,\HS) \\
\phi_3 &= \PROM(S_d,\DG)      &  \phi_6 &= \DEM(S_r,\PR) \\
&                               & \phi_7 &= \DEM(S_e,\PR)
\end{aligned}
\]
\end{example}

There is an interesting theoretical question regarding the interaction between right promotion/demotion and collision in AI deployment scenarios. In fact, one may argue that, when two fundamental rights $\Val{i}$ and $\Val{j}$ are promoted and demoted in the same deployment scenario $S$, then they collide in $S$:
\[
\PROM(S,\Val{i}) \wedge \DEM(S,\Val{j}) \to \collide{\Val{i}}{\Val{j}}{S}
\]
The principle above is not strictly necessary and can be debatable. Still, we may have good reasons to adopt it if your view regarding collision in AI deployment scenarios between fundamental rights is related to the conditions for their realization.

Finally, note that we may claim to assume monotonicity of right promotion or demotion: in other words, we can impose, e.g., that $\PROM (Y,\cval)$ is inconsistent with $\DEM (X,\cval)$ for any $X$ such that $Y$ implies $X$. Even this assumption is philosophically not needed, but it can be reasonably adopted.

\subsection{Priorities over Fundamental Rights and Right Adoption}\label{sec:priorities}
It is generally difficult to argue in abstract terms for a preference order over fundamental rights \cite{Alexy2014-ALECRA-2,Alexy2009}, because this kind of preference generally refers to the concept that some fundamental rights are considered more important or deserving of greater protection than others. The idea is less controversial if such a preference results from a complex balancing process applied to a multitude of AI deployment scenarios but is valid only in specific scenarios (see Section \ref{sec:balancing2}). Here, we simply provide some theoretical intuitions regarding priorities over fundamental rights.

Following an intuition in \cite{CalardoGR18}, rights can be ordered using complex constructions like the following:
\begin{gather*}
\Val{1}\otimes \Val{2} \otimes \dots \otimes \Val{n}
\end{gather*}
The interpretation of the expression $\Val{1}\otimes \Val{2} \otimes \dots \otimes \Val{n}$ is that $\Val{1}$ has a priority over $\Val{2}$, but, if $\Val{1}$ is demoted in the deployment scenario at hand, e.g. $S$, then $\Val{2}$ becomes preferred in that scenario, and so forth for the subsequent fundamental rights in the sequence.   

Once we have established any preference over fundamental rights in a given AI deployment scenario, and determined which rights are therein promoted or demoted, we can \emph{form a preference} for a specific right: if we write $\choice{X}{\Val{i}}$, it means that $\Val{i}$ is \emph{actually preferred and adopted} in the scenario described by $X$.\\

\begin{example}\label{ex:pandemic}
Consider a scenario $S$ where, to prevent the spreading of a lethal pandemic a provider develops an AI system that needs a massive but quick training and where we exceptionally use clinical data without the consent of  data owners. This means demoting privacy in favour of public health. Under this condition, we would have
\begin{gather*}
S=\mathit{pandemic}\wedge\neg \mathit{consent}\\
\mathbf{privacy}\otimes \mathbf{public\_health} \\
\DEM (X,\mathbf{privacy}) \\
\PROM (X,\mathbf{public\_health}).
\end{gather*}
\end{example}

\subsection{Defeasible Reasoning and Priorities over Fundamental Rights}\label{sec:balancing2}
The conceptual pillars presented in Sections \ref{sec:structure}-\ref{sec:priorities} pave the way for reconstructing how to reason about the AI impact towards fundamental rights. In particular, as we argued in Section \ref{sec:IPPC}, this type of legal reasoning amounts to a kind a balancing test. In Section \ref{sec:balancing_defeasible} we discussed the integration of \emph{definitional balancing} and \emph{defeasible reasoning} as a framework for addressing legal risks associated with AI.

\emph{Definitional balancing} is a method for resolving conflicts between competing rights by evaluating their importance and establishing general principles. It employs \emph{proportionality analysis} to weigh legitimate goals against rights limitations, ensuring fair and context-sensitive resolutions.
\emph{Defeasible reasoning} acknowledges the dynamic nature of legal decision-making, where conclusions may change based on new information or circumstances. This approach involves recognizing exceptions and prioritizing rules, allowing for flexibility in legal interpretations.

Together, these methods create a comprehensive framework for navigating conflicts among fundamental rights in AI deployment scenarios. 
More precisely, following the discussion in Section \ref{sec:balancing_defeasible}, the intuition is the following:\\

\begin{intuition}[Balancing Rights and Defeasibility for Legal Risk of AI]
Our approach is to reconstruct the process of balancing rights for risk assessment (as introduced in \cite{NovelliEtAl2023,NovelliEtAl2024,deon2025}) as follows:\\
\begin{enumerate}
 \item balancing rights for risk assessment first requires to run definitional balancing with respect to each specific hypothetical deployment scenario;
 \item defeasible reasoning is used to handle conflicts among deployment scenarios in regard to the impact of AI on rights and to aggregate the impact analysis to obtain an assessement with respect to the AI system or the GPAI.\\
\end{enumerate}
\end{intuition}

Let us use expressions such as 
\begin{gather*}
X \nmdash B \\
\end{gather*}
to denote that the sentence $B$ is a defeasible conclusion of $X$ (for the legal case, see \cite{handbook:deontic}). We may think of $\nmdash$ as any suitable  nonmonotonic consequence relation \cite{gab:foundations,sho:semantical,KraLehMag:nonmonotonic}. 

On account of the analysis developed in the previous sections, several reasoning patterns can be  identified. Let us consider some of them---the simplest ones---presented in a gentle and informal way\footnote{In a logical system, the following reasoning patterns correspond to inference rules. We present a version of these rules where, instead of considering expressions built with $\otimes$ of arbitrary length $n$, we focus only on expressions of the form $\Val{i} \otimes \Val{j} \otimes \Val{k}$.}.

Consider an AI deployment scenario $S$:

\begin{align}
\tag{Right adoption 1}
\text{IF } \qquad & S \nmdash \Val{i} \otimes \Val{j} \otimes \Val{k} \notag\\
\text{AND } \qquad & S \nmdash \neg \DEM (S,\Val{i})\notag \\
\text{THEN } \qquad & S \nmdash \choice{S}{\Val{i}} \notag
\end{align}

\begin{align}
\tag{Right adoption 2}
\text{IF } \qquad & S \nmdash \Val{i} \otimes \Val{j} \otimes \Val{k} \notag\\
\text{AND } \qquad & S \nmdash \DEM (S,\Val{i})\notag \\
\text{THEN } \qquad & S \nmdash \choice{S}{\Val{j}} \notag
\end{align}

\begin{align}
\tag{Right adoption 3}
\text{IF } \qquad & S \nmdash \Val{i} \otimes \Val{j} \otimes \Val{k} \notag\\
\text{AND } \qquad & S \nmdash \DEM (S,\Val{i})\notag \\
\text{AND } \qquad & S \nmdash \neg \collide{\Val{j}}{\Val{k}}{S}\notag \\
\text{THEN } \qquad & S \nmdash \choice{S}{\Val{k}} \notag
\end{align}

These reasoning patterns  allow us to defeasibly conclude \emph{actual preferences} in a given scenario $S$. The first two patterns essentially describe ways to adopt rights that are not demoted, while previously (and, in abstract terms, more preferred) rights are demoted. The third pattern states that any less preferred right $\Val{k}$, relative to a non-demoted right $\Val{j}$---which is adopted---is also adopted if it does not conflict with $\Val{j}$.\\

\begin{example}\label{ex:pandemic2}
Let us recall Example \ref{ex:pandemic} and elaborate it further. Indeed, in the light of the discussion in this section, we can reconstruct the case as follows:
\begin{gather*}
S\nmdash \mathbf{privacy}\otimes \mathbf{public\_health} \\
S\nmdash \neg\mathit{}\DEM (X,\mathbf{privacy}) \\
S\nmdash \PROM (X,\mathbf{public\_health})
\end{gather*}
Since $S$ demotes $\mathbf{privacy}$, then we can apply (Right adoption 2) and hold as true that $\choice{S}{\mathbf{public\_health}}$, i.e., we can conclude:
\[
S\nmdash \choice{S}{\mathbf{public\_health}}.
\]
\end{example}

\subsection{Deployment Scenarios and Fundamental Rights Analysis}\label{sec:conceptual}
Let us now build a procedure to articulate the different layers needed   to analyze the risk dimensions of AI systems and GPAIs. The methodology is detailed in two tables (Figure \ref{fig:AIsystems} and Figure \ref{fig:GPAI}).  Figure \ref{fig:AIsystems} outlines the steps for analyzing a specific high-risk AI system ($\sigma$),  while Figure \ref{fig:GPAI} extends this analysis to GPAIs ($\Sigma$), which possess a broader range of potential deployment scenarios.  The analysis proceeds in several steps:

\begin{enumerate}
    \item \textbf{Defining the Scope:} This involves identifying the AI system's purpose and the domain of its deployment ($\mathcal{D}_{\sigma}$ or $\mathcal{P}_{\Sigma}$). For GPAIs, this involves defining families of deployment scenario types.
    \item \textbf{Identifying Deployment Scenarios:} Specific scenarios ($S_i$) are defined within the deployment domain. For GPAIs, this is done for each deployment scenario type.
    \item \textbf{Describing Scenarios:}  Each scenario is characterized by its features ($f_1, \dots, f_m$).
    \item \textbf{Identifying Obligations and Rights:} The analysis identifies relevant legal obligations ($\OBL_{S_i}$) and fundamental rights ($\VAL_{S_i}$) implicated by each scenario.
    \item \textbf{Assessing Rights Impact:}  Each scenario is evaluated for its potential to promote or demote ($\PROM(S_i, \Val{j})$ or $\DEM(S_i, \Val{j})$) the identified rights.
    \item \textbf{Prioritizing Rights:}  Finally, a preference ordering among the affected rights is established where necessary to address potential conflicts.
   \item \textbf{Reasoning about Rights in Deployment Scenarios:} Finally, we can reason about the given deployment scenarios to determine which rights are protected (not demoted),
and which ones are demoted, and what is the optimal set
of deployment scenarios minimizing risk.\\
\end{enumerate}

By following these steps, the method allows for a comprehensive risk assessment, considering the potential for legal violations and fundamental rights infringements across a range of deployment contexts. The layered approach, from high-level scenarios to detailed applications, facilitates a nuanced understanding of the risks involved, supporting informed decision-making in AI governance and risk mitigation.  The approach explicitly recognizes that the set of deployment scenarios for GPAI systems is a superset of those for high-risk AI systems, reflecting the broader potential impact of GPAIs.


\begin{figure}[H]
\begin{center}
\begin{tabular}{||c |c|c||} 
 \hline
 {\bf Step} & {\bf What} & {\bf Notation} \\ [0.5ex] 
 \hline\hline
 1 & \thead{{\bf Purpose:}\\ Consider an AI system $\sigma$ and 
  establish the domain $\mathcal{D}_{\sigma}$ \\ of deployment for $\sigma$ (e.g., scheduling patients \\ visit 
  in e-health environments)} & $\mathcal{D}_{\sigma}$  \\
  \hline
 2 & \thead{{\bf Deployment Scenarios:} \\ Define the scenarios $S_1,\dots , S_n$ of $\mathcal{D}_{\sigma}$ \\ 
 (e.g., search for free slots, 
  retrieve a patient's appointments, \\
  book an appointment\dots)} & $\mathcal{D}_{\sigma} = \bigcup_{i=1}^n S_i$ \\
    \hline
 3 & \thead{{\bf Deployment Scenario Description:} \\ Describe each deployment scenario $S_i$ \\ in terms of its features $f_1, \dots , f_m$ \\ (e.g., book an appointment in terms of: \\
 patient's name, patient's disease,  patient's request, \dots)} & $S_i =\set{f_1, \dots , f_m}$ \\
 \hline
  4 & \thead{{\bf Obligations:} \\ Identify obligations $d_1, \dots , d_j$ relevant for each scenario $S_i$ \\ from legislation, codes of conduct, \dots \\
  (e.g., the obligation for the deployer \\ to ask for the consent of the affects individual \\ to the processing of his or her personal data)} & $\OBL_{S_i} =\set{d_1, \dots , d_j}$ \\
 \hline
  5 & \thead{{\bf Fundamental  Rights:} \\ 
  Given a finite set of basic rights $R=\set{\val{1}, \dots , \val{k}}$, \\ identify the set of relevant fundamental rights 
  $\set{\Val{1}, \dots , \Val{n}}$
  \\ for each scenario $S_i$ from legislation, codes of conduct, \dots \\
  (e.g., privacy, health, participation, transparency, \\nondiscrimination, equity, and accountability);\\
  each $\Val{j}$, $1\leq j \leq n$, is a Boolean combination \\ of basic rights from $R$ (and, possibly, also of obligations)} 
   & $\VAL_{S_i} =\set{\Val{1}, \dots , \Val{n}}$ \\
 \hline
  6 & \thead{{\bf Rights Promotion/Demotion:} \\ 
  Establish for each scenario \\ $S_i =\set{f_1, \dots , f_m}$ \\ which rights in $\VAL_{S_i}$ are promoted or demoted by $S_i$ \\
  (e.g., 
  when encryption is not used,
  then privacy is demoted)} & \thead{$\forall \Val{j}\in \VAL_{S_i}$ either \\
  $\PROM (S_i,\Val{j})$ or\\
  $\DEM (S_i,\Val{j})$ or \\
  undefined.
  } \\
 \hline
  7 & \thead{{\bf Rights Preference:} \\ 
  Establish for each scenario \\ $S_i =\set{f_1, \dots , f_m}$ \\ if there is a preference ordering on some or all rights in $\VAL_{S_i}$ \\
  (e.g., with life-threatening health is more important \\ than data protection)} & \thead{$
  f_1\wedge \dots \wedge f_m \nmdash 
  \Val{1}\otimes \dots \otimes  \Val{n}$ \\
  where \\ $\set{\Val{1}, \dots , \Val{n}}\subseteq \VAL_{S_i}$
  } \\
  \hline
8 & \thead{{\bf Reasoning about Rights}} & \thead{See Sections \ref{sec:balancing2} and \ref{sec:heuristics}}
\\ [1ex] 
 \hline
\end{tabular}
    \caption{Deployment scenario analysis for high-risk AI systems}
  \label{fig:AIsystems}
  \end{center}
\end{figure} 
\begin{figure}[H]
\begin{center}
\begin{tabular}{||c |c|c||} 
 \hline
 {\bf Step} & {\bf What} & {\bf Notation} \\ [0.5ex] 
 \hline\hline
 1 & \thead{{\bf General Purpose:}\\ Consider a GPAI $\Sigma$ and 
  establish its general purpose as $\mathcal{P}_{\Sigma}$ \\  (e.g., an LLM for generating legal texts)} & $\mathcal{P}_{\Sigma}$  \\
  \hline
 2 & \thead{{\bf Family of Deployment Scenario Types:} \\ Define for $\mathcal{P}_{\Sigma}$ a family $\mathcal{F}_{\mathcal{P}_{\Sigma}}$ of deployment scenario types $\mathcal{D}^1_{\Sigma}, \dots , \mathcal{D}^m_{\Sigma}$ \\
 (e.g., judicial decision-making, legislative legal drafting, \dots)} & $\mathcal{F}_{\mathcal{P}_{\Sigma}} = \set{\mathcal{D}^i | 1\leq i \leq n}$ \\
    \hline
    3 & \thead{{\bf Deployment Scenarios:} \\ Define for each deployment scenario type $\mathcal{D}^i_{\Sigma}$ in $\mathcal{F}_{\mathcal{P}_{\Sigma}}$ \\
   a set of deployment scenarios $S_1,\dots , S_n$ of $\mathcal{D}^i_{\Sigma}$\\
   (e.g., for judicial decision-making: summarization, legal case retrieval, \\ legal judgment prediction,\dots)} & $\mathcal{D}^i_{\Sigma} = \bigcup_{j=1}^n S_j$ \\
    \hline
 4 & \thead{{\bf Scenario Description:} \\ see Figure \ref{fig:AIsystems}} & $S_i =\set{f_1, \dots , f_m}$ \\
 \hline
  5 & \thead{{\bf Obligations:} \\ see Figure \ref{fig:AIsystems}} & $\OBL_{S_i} =\set{d_1, \dots , d_j}$ \\
 \hline
  6 & \thead{{\bf Fundamental  Rights:} \\ 
  see Figure \ref{fig:AIsystems}}
   & $\VAL_{S_i} =\set{\Val{1}, \dots , \Val{n}}$ \\
 \hline
  6 & \thead{{\bf Rights Promotion/Demotion:} \\ 
  see Figure \ref{fig:AIsystems}} & \thead{$\forall \cVal_j\in \VAL_{S_i}$ either \\
  $\PROM (S_i,\Val{j})$ or\\
  $\DEM (S_i,\Val{j})$ or \\
  undefined.
  } \\
 \hline
  7 & \thead{{\bf Rights Preference:} \\ 
  see Figure \ref{fig:AIsystems}} & \thead{$
  f_1\wedge \dots \wedge f_m \nmdash 
  \Val{1}\otimes \dots \otimes  \Val{n}$ \\
  where \\ $\set{\Val{1}, \dots , \Val{n}}\subseteq \VAL_{S_i}$
  }
  \\
   \hline
8 & \thead{{\bf Reasoning about Rights}} & \thead{See Sections \ref{sec:balancing2} and \ref{sec:heuristics}}
\\ [1ex] 
 \hline
\end{tabular}
    \caption{Deployment scenario analysis for GPAIs}
  \label{fig:GPAI}
  \end{center}
\end{figure}

\subsection{Definitional Balancing to Minimize the Legal Risk of AI}\label{sec:heuristics}
We use the qualitative framework from the previous sections to formulate some heuristics to minimize the legal risk of an AI system or GPAI: what we present here is in fact a kind of \emph{optimization procedure} (in the sense of \cite{Alexy2014-ALECRA-2,Alexy2009}) with respect to the fundamental rights involved in the deployment scenarios of a given AI. \\

\begin{definition}[AI Systems: Right Impact]\label{def:optimisation}
Let $\mathcal{D}_{\sigma} = \bigcup_{i=1}^n S_i$ be the application domain of an AI system $\sigma$. 
By convention, if a right $\cVal$ occurs in any $n$th position in an $\otimes$-expression of length $y$, then we indicate it by adding $\langle n,y\rangle$ as a subscript to $\cVal$. For example, we might write $\cVal_{\langle1,y\rangle} \,\otimes\, \cVal_{\langle2,y\rangle} \,\otimes\, \dots \,\otimes\, \cVal_{\langle y,y\rangle}$.

We define for $\mathcal{D}_{\sigma}$ \emph{the degree of right impact as $\mathbf{Degree}(\mathcal{D}_{\sigma})=\Xi - \Delta$} where 
\begin{gather*}
   \Xi=\left[\sum_{\forall S_i, \forall \cVal_{\langle x, y  \rangle} : S_i\nmdash 
   \choice{S_i}{\cVal{\langle x, y  \rangle}}}\left(
   \frac{y}
    {x}\right)\right]
\end{gather*}
%
%
\begin{gather*}
   \Delta=\left[\sum_{\forall S_i \forall \cVal_{\langle x, y  \rangle} : S_i\nmdash \DEM(S_i, \cVal_{\langle x, y  \rangle})}\left(
   \frac{y}
    {x}\right)\right]
\end{gather*}
\end{definition}

\begin{definition}[AI System: Legal Risk Minimization]\label{def:minimization}
The legal risk of an AI system $\sigma$ is minimal iff $\sigma$ is designed for a set $\mathcal{D}'_{\sigma}=\bigcup_{i=1}^m S_i$ of deployment scenarios, where $\mathcal{D}'_{\sigma}\subseteq \mathcal{D}_{\sigma}$, such that for all $X\in (2^{\bigcup_{i=1}^p S_i})-\set{\emptyset}$ where $X\not= \bigcup_{i=1}^m S_i$, we have that $\mathbf{Degree}(\mathcal{D}'_{\sigma})\geq \mathbf{Degree}(X)$.\\
\end{definition}
%
Risk minimization 
amounts to searching in $\mathcal{D}_{\sigma}$ the subset of deployment scenarios which demotes as little as possible the best-preferred rights. According to Table \ref{fig:GPAI}, moving from AI systems to GPAI is rather straightforward.\\
\begin{definition}[AI Systems: Right Impact]\label{def:optimisation}
Let $\Sigma$ be a GPAI with the general purpose $\mathcal{P}_{\Sigma}$. Define for $\mathcal{P}_{\Sigma}$ a family $\mathcal{F}_{\mathcal{P}_{\Sigma}}$ of deployment scenario types $\mathcal{D}^1_{\Sigma}, \dots , \mathcal{D}^m_{\Sigma}$, being each $\mathcal{D}^j_{\Sigma} = \bigcup_{i=1}^n S_i$ an application domain for $\Sigma$.

We define for $\mathcal{P}_{\Sigma}$ \emph{the degree of right impact $\mathbf{Degree}(\mathcal{P}_{\Sigma})$} as $\sum^{m}_{i=1} \mathbf{Degree}(\mathcal{D}^i_{\Sigma})$.\\
%
\end{definition}

\begin{definition}[GPAI: Legal Risk Minimization]\label{def:minimization}
The legal risk of a GPAI $\Sigma$ is minimal iff $\Sigma$ is designed for a set $\mathcal{P}'_{\Sigma}=\bigcup_{i=1}^m \mathcal{D}^i_{\Sigma}$ of domains, where $\mathcal{P}'_{\Sigma}\subseteq \mathcal{P}_{\Sigma}$, such that for all $X\in (2^{\bigcup_{i=1}^p \mathcal{D}^i_{\Sigma}})-\set{\emptyset}$ where $X\not= \bigcup_{i=1}^m \mathcal{D}^i_{\Sigma}$, we have that $\mathbf{Degree}(\mathcal{P}'_{\Sigma})\geq \mathbf{Degree}(X)$.
\end{definition}

\section{Conclusions}
This chapter has presented a novel conceptual framework for qualitative risk assessment of AI, focusing on the challenges posed by the EU AI Act and the need for a nuanced understanding of legal risk.  We moved beyond simplistic risk categorization, advocating for a more granular analysis that considers the interplay between various risk factors and their impact on fundamental rights.

Our framework leverages the strengths of both definitional balancing and defeasible reasoning. Definitional balancing provides a structured approach to resolving conflicts between competing rights and interests, incorporating proportionality analysis to ensure fair and context-sensitive outcomes.  Defeasible reasoning, on the other hand, acknowledges the dynamic nature of legal decision-making, allowing for the revision of conclusions in light of new information and changing circumstances.  The combination of these two methodologies yields a flexible yet robust framework for evaluating AI risks.

The core of our approach lies in the detailed analysis of deployment scenarios.  We proposed a layered approach, progressing from high-level scenarios to specific applications, enabling a more precise identification of potential legal violations and impacts on fundamental rights.  This layered analysis is particularly crucial for General Purpose AI systems, whose broad applicability necessitates a comprehensive assessment across a wider range of potential uses.

We introduced some general principles to represent the interactions between deployment scenarios and fundamental rights.  These principles allow for a structured representation of the complex relationships involved in AI risk assessment, providing a basis for more rigorous analysis.  Furthermore, we claim that defeasible reasoning must incorporate operators for expressing when rights are promoted and demoted, providing a mechanism for capturing the dynamic interplay between competing rights within specific contexts. 

While this chapter presents a high-level conceptual framework, future work will focus on developing a more complete formal model.  This will involve a detailed exploration of the suitable logical systems and semantics, a more thorough investigation of the properties of the proposed reasoning patterns, and the development of effective algorithms for automated risk assessment.  The integration of these elements will lead to a practical tool for assisting regulators, developers, and users in navigating the complex legal and ethical landscape of AI.  Our ongoing research aims to bridge the gap between theoretical frameworks and practical applications, thereby contributing to the responsible development and deployment of AI technologies.

\subsection*{Acknowledgements}
Antonino Rotolo was  supported by the projects EUSAiR “EU Regulatory Sandboxes for AI” (DIGITAL-2024-AI-ACT-06-SANDBOX), CN1 “National Centre for HPC, Big Data and Quantum Computing” (CUP: J33C22001170001), and PE01 “Future Artificial Intelligence Research” FAIR (CUP: J33C22002830006). Jose Miguel Angel Garcia Godinez was  supported by the project CN1 “National Centre for HPC, Big Data and Quantum Computing” (CUP: J33C22001170001). Giovanni Sartor was supported by the projects H2020 ERC Project CompuLaw (G.A. 833647) and PE01 “Future Artificial Intelligence Research” FAIR (CUP: J33C22002830006).

\bibstyle{apalike}
\bibliography{finalbiblio}

\end{document}